# From Scope to Script: An Automated Report Generation Model for Gastrointestinal Endoscopy

Evandros Kaklamanos[1], Kristjana Kristinsdottir[1], Jonathan Huang[1], Dustin Carlson[2], Rajesh Keswani[3], John Pandolfino[2], Mozziyar Etemadi[1,4]

[1] Research & Development, Department of Information Services, Northwestern Medicine, Chicago, Illinois, USA
[2] Kenneth C. Griffin Esophageal Center of Northwestern Medicine, Department of Medicine, Division of Gastroenterology and Hepatology, Feinberg School of Medicine, Northwestern University, Chicago, Illinois, USA
[3] Division of Gastroenterology and Hepatology, Department of Medicine, Feinberg School of Medicine, Northwestern University, Chicago, Illinois, USA
[4] Department of Anesthesiology, Feinberg School of Medicine, Northwestern University, Chicago, Illinois, USA

evandros.kaklamanos@nm.org

**Abstract.** Endoscopic procedures such as esophagogastroduodenoscopy (EGD) and colonoscopy play a critical role in diagnosing and managing gastrointestinal (GI) disorders. However, the documentation burden associated with these procedures place significant strain on gastroenterologists, contributing to inefficiencies in clinical workflows and physician burnout. To address this challenge, we propose a novel automated report generation model that leverages a transformer-based vision encoder and text decoder within a two-stage training framework. In the first stage, both components are pre-trained on image/text caption pairs to capture generalized vision-language features, followed by fine-tuning on images/report pairs to generate clinically meaningful findings. Our approach not only streamlines the documentation process but also holds promise for reducing physician workload and improving patient care.

## 1 Introduction

GI upper endoscopy and colonoscopy are critical for diagnosing and treating conditions like inflammatory bowel disease, gastroesophageal reflux disease, and colorectal cancer [1, 2, 3]. Historically, these procedures were documented using manual free-text entries. This approach afforded clinicians the flexibility to capture nuanced findings and patient-specific details. However, this process is not only time-consuming and prone to variability but also heavily dependent on clinician expertise, often leading to inconsistencies in documenting critical findings—such as lesion localization, bowel preparation quality, and procedural nuances [4, 5].

To address these limitations, templated systems were developed. These systems integrate predefined prompts and structured data entry field—often based on standards set by organizations such as the European and American Societies of Gastrointestinal Endoscopy (ESGE and ASGE)—to promote consistency and accuracy in report content [5, 6, 7]. While templated reporting has helped to mitigate these inconsistencies inherent in free-text documentation, its rigid structure can restrict the ability to capture atypical or complex clinical scenarios that require individualized narrative descriptions. Furthermore, despite their intent to streamline workflows, templated systems still impose a documentation burden, requiring clinicians to manually navigate these templates. The increasing volume and complexity of endoscopic procedures, coupled with this administrative burden of detailed documentation, has contributed to physician burnout, especially in physicians with less experience [8].

In contrast, automated medical report generation has seen considerable progress in radiology. Transformer-based architectures and attention mechanisms have demonstrated remarkable success in mapping radiological images to textual descriptions by leveraging large-scale multimodal datasets [9, 10]. Vision-language models, which typically integrate vision transformers (ViTs) with large language models (LLMs), employ cross-modal attention to align visual features [11, 12]—such as lesion localization in X-rays—with detailed clinical narratives. These systems not only adhere to established radiological lexicons but also incorporate patient-specific findings, offering a more dynamic alternative to traditional templated reports. However, translating these successes to GI procedures presents unique challenges. The dynamic nature of endoscopic video data, the need for fine-grained lesion classification, and the detailed documentation of procedural steps (e.g., cecal intubation time) require a tailored approach. Prior research in GI AI has predominantly focused on narrow tasks such as surgical tool localization [13, 14], polyp detection [15, 16], or anatomical landmark identification [16].

Recent advancements highlight promising directions in gastroenterology. For instance, deep learning systems like the Image Recognition-Based Structured Report Generation System (ISRGS) utilize traditional models such as Convolutional Neural Networks (CNNs) to combine the output of several submodules including lesion detection, site location, and lesion classification to produce the final templated report [17]. Similarly, AI-EARS, developed a similar approach combining the output of multiple specialized CNNs to produce a report [18]. However, these systems rely on rigid templates and lack adaptability to nuanced clinical contexts.

Despite these advances, progress in AI-driven GI report generation lags radiology, where public datasets like MIMIC-CXR [19] have catalyzed innovation. Endoscopy faces unique challenges: endoscopic procedures generate long video streams that demand prohibitive storage and computational resources for large-scale curation. These videos exhibit high variability in resolution, lighting, and frame rates due to differences in endoscope hardware. Furthermore, endoscopic videos are not routinely archived in clinical workflows [20]—unlike radiology images, which are systematically stored as part of standard of care. Clinicians prioritize real-time decision-making during procedures over video recording, leaving sparse, inconsistently annotated data for retrospective analysis. These barriers collectively stifle the creation of public datasets, limiting the development of robust, generalizable vision-language models for endoscopy.

In this paper, we propose a novel transformer-based framework designed specifically for automated report generation in GI endoscopy compatible with both templated reporting and free-text systems. Our approach leverages a two-stage training process that first pre-trains a vision encoder and text decoder on extensive image-text caption pairs and subsequently fine-tunes them on image-report pairs. By addressing the unique challenges of GI procedures, our method aims to reduce documentation variability and clinician workload while enhancing the accuracy of endoscopic reports.

## 2 Methods

### 2.1 Data Preparation

For stage 1, we curated a fully de-identified internal multimodal dataset of 775,990 image-text pairs from screenshots acquired during 32,949 EGD and 72,314 colonoscopy procedures. Table 1 summarizes the dataset across the training, validation, and test split. Procedural screenshots were taken as part of standard-of-care, capturing key anatomical landmarks and any relevant findings. Images were resized to 224×224 pixels and normalized using ImageNet statistics. Accompanying text captions were tokenized using the GPT-2 byte-pair encoding (BPE) tokenizer [21]. The procedures and screenshots were annotated post-operatively by the performing physicians using a structured reporting interface. Physicians populated mandatory fields via dropdown menus, including anatomical regions, findings, and optional narrative fields allowing for more nuanced observations.

**Table 1.** Composition and stratification of endoscopic dataset.

| Dataset | Patients | Patient encounters | Procedures | EGDs | Colonoscopies | Images |
|---|---|---|---|---|---|---|
| Train | 78,537 | 85,512 | 102,188 | 32,099 | 70,089 | 754,124 |
| Validation | 1,289 | 1,292 | 1,536 | 429 | 1,107 | 10,762 |
| Test | 1,289 | 1,292 | 1,539 | 421 | 1,118 | 11,104 |

Stage 2 utilized the same de-identified patient, procedures, and screenshots but instead paired the detailed findings section of the operative reports with all the screenshots captured during the procedure. The image annotations acquired in stage 1 were not used as input to the model for generating the report in stage 2. The same structured reporting interface was used for documenting the findings section, allowing for both free-text and templated options. To mitigate computational bottlenecks, procedures that captured more than 12 screenshots were excluded for training and evaluation due to GPU memory constraints. A patient-level split was applied ensuring there was no patient or procedure overlap between the train, validation or test set. The train/validation/test split remained consistent across both stages. Figure 1 describes the data acquisition process as part of the physician's workflow.

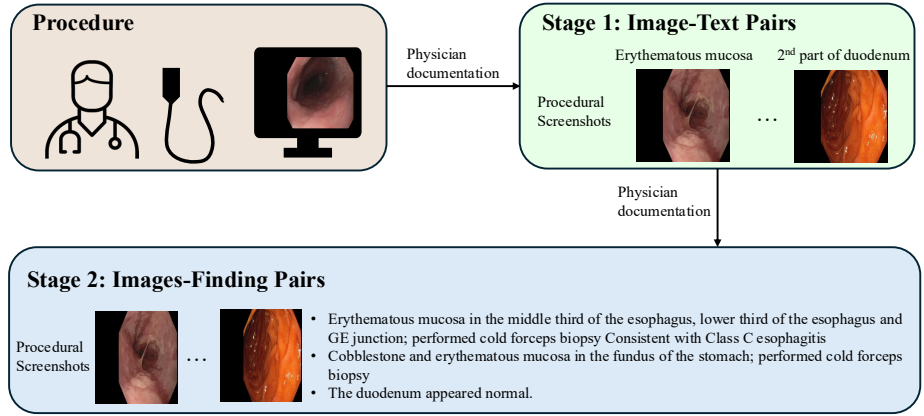

**Fig. 1.** Data acquisition for stage 1 and stage 2 as part of physician workflow.

### 2.2  Model Architecture

The model is based on a transformer architecture that links a vision encoder and text decoder through cross-attention (see Fig. 2). The vision encoder is initialized with SigLIP (B/16) [22] and the text decoder is a randomly initialized transformer decoder with 6 layers and 6 attention heads per layer. The text input begins with a special [BOS] token and is decoded auto-regressively until the maximum number of steps is reached or generating the [EOS] special token. The model is trained in two stages: (i) pre-training on image-text caption pairs and (ii) fine-tuning for generating the findings section from multiple procedural images.

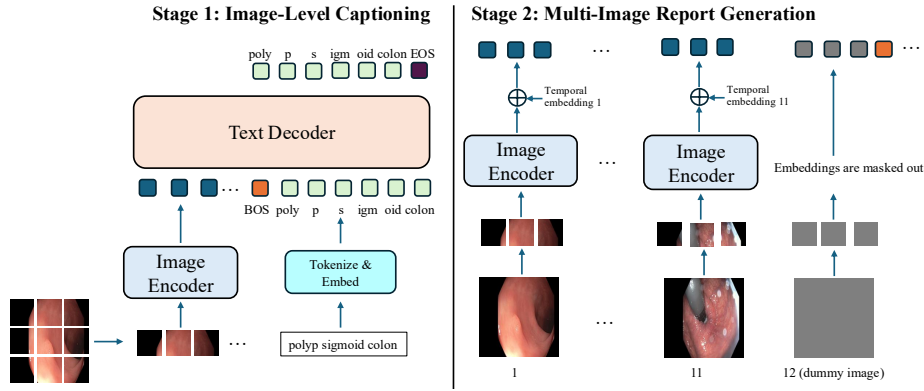

**Fig. 2.** Model architecture overview. Procedures with less than 12 images have "dummy" images where their embeddings are masked out for the text decoder.

**Stage 1: Image-Level Captioning.** An input endoscopic image, $X \in \mathbb{R}^{224 \times 224 \times 3}$, is split into $N = 196$ non-overlapping $16 \times 16$ patches. The vision encoder $E(\cdot)$ maps the image X to a latent representation $z = E(X) \in \mathbb{R}^d$ ($d=768$). During pre-training, a paired caption $y = (y_1, y_2, ... y_T)$ is available, and the text decoder $D(\cdot)$ generates a probability over the vocabulary conditioned on the encoder output. The language modeling (LM) loss is the cross-entropy (CE) loss applied as the following without any label smoothing

$$\mathcal{L}_{LM} = \frac{1}{N+1}\sum_{i=1}^{N+1} \text{CE}\left(y_i, p(y_i|X, \{y_j, j = 0, ..., i = 1\})\right) \quad (1)$$

**Stage 2: Multi-Image Report Generation.** In the subsequent stage, the pre-trained vision encoder $E(\cdot)$ and text decoder $D(\cdot)$ remain unfrozen and are jointly fine-tuned on the dataset consisting of multiple images paired with clinical findings. For a given sequence of $N \leq 12$ images $\{X_1, X_2, ..., X_N\}$, each image is processed individually by the vision encoder to yield corresponding embeddings $\{z_1, z_2, ..., z_{N \times 196}\}$. A context mask is applied such that procedures with less than 12 images have their corresponding embeddings masked out preventing them from attending to other image embeddings. Afterwards, a learned temporal embedding is added, and the embeddings from each image are concatenated. Again, the model is optimized over the language modeling loss.

## 2.3   Training Setup

The model was trained with PyTorch on a system equipped with 2 NVIDIA Titan RTX GPUs using mixed precision (fp16). For stage 1, the training was conducted over 10 epochs with a nominal batch size of 12. To effectively simulate a larger batch size and stabilize gradient estimates, gradient accumulation was used over 32 iterations, resulting in an effective batch size of 768 samples per parameter update. Stage 2 was trained over 30 epochs with a nominal batch size of 1 and gradient accumulation over 32 iterations. For both stages, optimization was performed using the Adam optimizer with a peak learning rate of $6 \times 10^{-4}$. The learning rate schedule included a linear warmup for the first 5% of iterations, followed by cosine annealing, during which the learning rate was reduced to 10% of its peak value.

## 3   Results and Discussion

### 3.1   Automatic Text Generation Performance

We evaluated our model using standard natural language generation (NLG) metrics: BLEU [23], METEOR [24], and ROUGE [25]. Model-generated reports were generated using greedy decoding to ensure deterministic outputs and compared to the gastroenterologist-documented reports for each procedure [26]. Overall, there was a high level of concordance between model-generated and ground truth reports, highlighting the potential of this model to produce reports which could be efficiently edited by a gastroenterologist to ensure clinical accuracy. Detailed results of this evaluation are presented in Table 2.

Table 2. Evaluation of model-generated reports on physician findings on the test set.

| Dataset | BLEU-1 | BLEU-2 | BLEU-3 | BLEU-4 | METEOR | ROUGE |
|---|---|---|---|---|---|---|
| Test | 0.590 | 0.445 | 0.371 | 0.324 | 0.520 | 0.546 |

**Impact of Vision-Language Pre-Training.** To investigate the impact of image-caption pre-training, we conducted an ablation study with (stage 1) and without pre-training the vision encoder and text decoder. The vision encoder was initialized with SigLIP (B/16) and the text decoder was randomly initialized for the experiment with no pre-training. The absence of pre-training led to a significant decrease of 72.96% in the ROUGE score, highlighting the effectiveness of image-caption pre-training (Table 3).

Table 3. NLG metrics on the validation set with and without image-caption pre-training.

| Experiment | BLEU-1 | BLEU-2 | BLEU-3 | BLEU-4 | METEOR | ROUGE |
|---|---|---|---|---|---|---|
| No pretraining | 0.334 | 0.192 | 0.136 | 0.109 | 0.315 | 0.318 |
| Pre-training | 0.595 | 0.448 | 0.372 | 0.325 | 0.525 | 0.550 |

### 3.2 Cross-Attention Grounding Tokenization Constraints

Figure 3 demonstrates the model's ability to generalize to unseen cases using a test set example where the generated caption accurately matches the ground truth label (e.g., "polyp rectum"). The last-layer cross-attention maps (averaged across heads) reveal that the vision encoder successfully localizes the polyp to the correct anatomical region, with the highest attention weight (brightest yellow) aligning precisely with the lesion site. This alignment between model attention and clinical correspondence between vision and text ground truth provides confidence that the model effectively learns clinically relevant features and provides an avenue for increased model interpretability.

However, this example reveals a key weakness: the generic tokenizer splits "polyp" into sub-word units ("poly" and "p"), leading to disjointed attention patterns for the fragmented tokens. While the model still associates both sub-words with the polyp region, this fragmentation complicates attention map interpretation and may degrade performance by yielding less optimal sequence lengths. A medical domain specific tokenizer trained on GI lexicons (e.g. merging "poly" and "p" into a single "polyp" token) would consolidate both text generation and attention visualization, ensuring that clinically meaningful concepts map to unified visual regions.

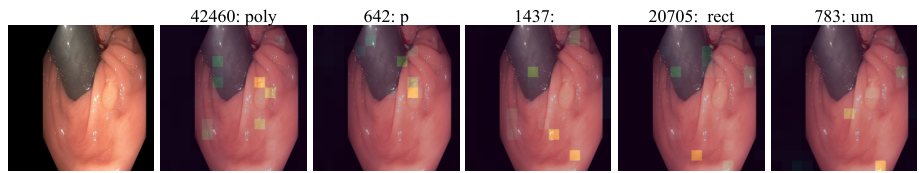

**Fig. 3.** Cross-attention guides anatomically grounded text generation. Left to right: original endoscopic image; overlayed cross-attention maps for predicted tokens with token id and decoded sub-word as sub-title.

### 3.3 Case Studies: Model Generated Reports vs Physician Reports

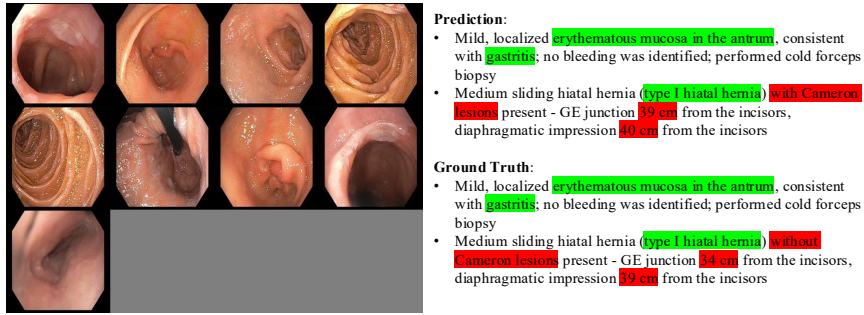

**Fig. 4.** Discrepancies in measurement reporting: Model vs. physician analysis. Red text indicates error in model prediction. Green text indicates correctly identified finding with model.

Figure 4 compares the model's report to the physician's ground truth for a case of sliding hiatal hernia and gastritis, both correctly identified by the model. However, the model misreports the GE junction distance from the incisors and the diaphragmatic impression distance from the incisors. These errors highlight a critical limitation: the model's inability to infer precise spatial measurements from static screenshots. Unlike categorical findings (e.g. gastritis), which rely on visual-semantic correlations, metric reporting requires explicit spatial calibration tied to endoscopic tools or anatomical reference points. This gap highlights the need for a specialized measurement module in the architecture, such as a depth-aware encoder or coordinate regression head, to map pixel distances to real-world metrics.

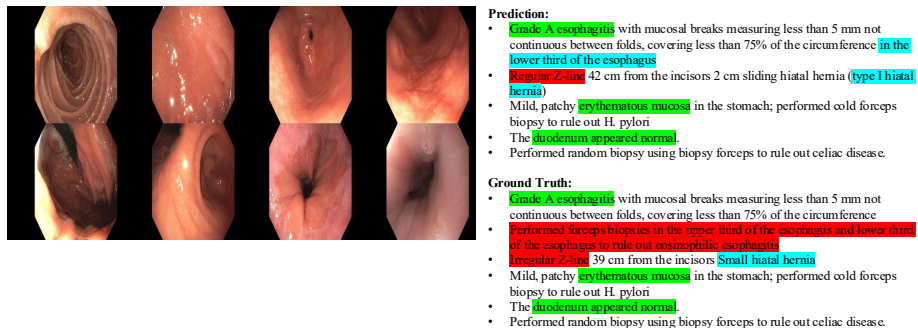

**Fig. 5.** Reporting protocol adherence: Model vs. physician. Red text indicates error in model prediction. Green text indicates correctly identified finding with model. Blue text indicates correctly identified finding with model but not by physician.

Figure 5 demonstrates the model's ability to improve the diagnostic granularity while overlooking procedural details. The physician report describes "grade A esophagitis" and "small hiatal hernia", whereas the model specifies "grade A esophagitis in the lower third of the esophagus" and "sliding hiatal hernia (type I hiatal hernia)". However, the model fails to note that biopsies were taken in the upper third and lower third of the esophagus, a detail included in the physician's report. This omission stems from two limitations: (1) biopsies are procedural actions often not visually apparent in static images, and (2) the model's input is wholly dependent on physician-selected frames. If a screenshot of the biopsy is not captured, the model cannot infer its occurrence.

This example exposes a broader challenge: static images cannot encapsulate the temporal workflow of endoscopy. Biopsies, tool interactions, and dynamic findings (e.g. esophageal spasms) require video context to document fully. Future work should adopt video-language modeling to encode temporal context, such as tool usage or sequential anatomical exploration. For instance, a video transformer could track biopsy events across frames, enabling the decoder to report both findings and interventions.

## 4   Conclusion

Manual and templated endoscopic reporting remain prone to variability, inefficiency, and rigidity. We present the first transformer-based framework that automates report generation using a two-stage training strategy, combining image-text pretraining with endoscopic report fine-tuning. The model can generate coherent and accurate findings based on procedural screenshots. Nevertheless, there are limitations such as relying on static screenshots, which impairs measurement precision and procedural context. To overcome this, future research should incorporate video-language modeling and depth-aware encoders. As the first end-to-end GI report generation transformer model, our approach shows promise in reducing documentation burden while preserving clinical nuance.

**Acknowledgements.** This preprint has not undergone peer review (when applicable) or any post-submission improvements or corrections. The Version of Record of this contribution is published in Deep Generative Models, and is available online at https://doi.org/10.1007/978-3-032-05472-2_13.

**Disclosure of Interests.** R.K Boston Scientific (Consulting), Medtronic (Consulting). D.C. Medtronic (Speaking, Consulting, License); Diversatek (Consulting); Braintree (Consulting); Medpace (Consulting); Phathom Pharmaceuticals (Speaking; Consulting); Regeneron/Sanofi (Speaking). J.P. Sandhill Scientific/Diversatek (Consulting, Grant), Takeda (Speaking), Astra Zeneca (Speaking), Medtronic (Speaking, Consulting, Patent, License), Torax/Ethicon (Speaking, Consulting), EndoGastric Solutions (Advisory Board), Phathom (Speaking, Consulting). Other authors have no conflicts to disclose.